\documentclass{paper}
\usepackage{amsmath,amsthm,mleftright}
\usepackage{amssymb,mathtools}
\DeclareMathAlphabet\mathcal{OMS}{cmsy}{m}{n}
\DeclareMathAlphabet\mathbfcal{OMS}{cmsy}{b}{n}
\usepackage{tikz-cd}
\usepackage{xspace}
\usepackage{makecell}
\usepackage{graphicx,booktabs}
\usepackage{ragged2e}
\usepackage{booktabs,multirow,tabularx}
\usepackage{paralist}
\usepackage{hyperref}
\hypersetup{%
    pdfmenubar=true,       
    pdffitwindow=false,     
    pdfstartview={FitH},    
    pdftitle={Image Reconstruction for Multispectral CT},
    colorlinks=true,       
    linkcolor=red,          
    citecolor=red,        
    filecolor=magenta,      
    urlcolor=cyan,           
    pdfborder = {0,0,0}
}
\usepackage{acro}
\usepackage{cleveref}
\usepackage{algorithm}
\usepackage[noend]{algpseudocode}
\makeatletter
\def\BState{\State\hskip-\ALG@thistlm}
\makeatother
\Crefname{equation}{}{}
\crefname{equation}{}{}
\usepackage[title]{appendix}
\usepackage{todonotes}
\usepackage{inputenc}
\theoremstyle{plain}
 
 \newtheorem{proposition}{Proposition}

\theoremstyle{definition}

\theoremstyle{remark}

\begin{document}

\title{Fracture interactive geodesic active contours for bone segmentation}
\author{Liheng Wang\thanks{State Key Laboratory of Mathematical Sciences, Academy of Mathematics and Systems Science, Chinese Academy of Sciences, Beijing 100190, China; University of Chinese Academy of Sciences, Beijing 100190, China.},  
Licheng Zhang\thanks{Department of Orthopedics, The Fourth Medical Center of Chinese PLA General Hospital, Beijing 100048, China; National Clinical Research Center for Orthopedics, Sports Medicine and Rehabilitation, Beijing 100048, China.},   
Hailin Xu\thanks{Department of Trauma and Orthopedics, People's Hospital Peking University, Beijing 100044, China.}, 
Jingxin Zhao\footnotemark[2],\\ 
Xiuyun Su\footnotemark[2], 
Jiantao Li\footnotemark[2], 
Miutian Tang\footnotemark[3], 
Weilu Gao\footnotemark[2], \\ 
Chong Chen\footnotemark[1]
}

\maketitle

\begin{abstract}
For bone segmentation, the classical geodesic active contour model is usually limited by its indiscriminate feature extraction, and then struggles to handle the phenomena of edge obstruction, edge leakage and bone fracture. Thus, we propose a fracture interactive geodesic active contour algorithm tailored for bone segmentation, which can better capture bone features and perform robustly to the presence of bone fractures and soft tissues. Inspired by orthopedic knowledge, we construct a novel edge-detector function that combines the intensity and gradient norm, which guides the contour towards bone edges without being obstructed by other soft tissues and therefore reduces mis-segmentation. Furthermore, distance information, where fracture prompts can be embedded, is introduced into the contour evolution as an adaptive step size to stabilize the evolution and help the contour stop at bone edges and fractures. This embedding provides a way to interact with bone fractures and improves the accuracy in the fracture regions. Experiments in pelvic and ankle segmentation demonstrate the effectiveness on addressing the aforementioned problems and show an accurate, stable and consistent performance, indicating a broader application in other bone anatomies. Our algorithm also provides insights into combining the domain knowledge and deep neural networks.
\end{abstract}


\section{Introduction}
\label{Section 1}

Because of the ability to provide high contrast and clear structures, computed tomography (CT) has been a crucial imaging tool in many orthopedic anatomies, such as the pelvis and the ankle. Accurate segmentation of bone CT images serves as a direct and critical reference throughout trauma diagnosis, preoperative planning, surgical reduction and postoperative evaluation \cite{han2021fracture}. As far as we know, in most cases, clinical annotation of bones still relies on manual assistance, typically facilitated by Mimics\footnote{https://en.wikipedia.org/wiki/Mimics.}, a medical image processing software equipped with simple thresholding and region growing algorithms. This usually entails massive interactions, such as threshold tuning and selection of seed points, followed by manual mask splitting and merging to obtain the bone regions. Moreover, these algorithms usually yield spongy bone masks in the case without additional manual intervention, thereby potentially limiting their further usage. 

In the last decade, deep learning based algorithms have attracted increasing attention and achieved great success in medical image segmentation. Among these, U-Net \cite{UNET} is an empirically effective and widely recognized convolutional network, which features a downsampling encoder, an upsampling decoder, and skip connections that facilitate direct information transfer from the encoder to the decoder. Research efforts have been focused on enhancing U-Net from different aspects, such as extending to 3D version \cite{3DUNET,VNET}, designing task-specific self-configuration \cite{NNUNET}, and improving the backbone with residual connection \cite{RESUNET,li2018h} or Transformer \cite{Cao2021SwinUnetUP}. In addition, U-Net and its variants have also been applied in many anatomies, such as pelvis \cite{DEEP_LEARNING_TO_SEGMENT_PELVIC,PELVIC_FRACTURE_SEGMENTATION}, retinal vessel \cite{9413346,9815506}, and breast tumor \cite{CHEN2023109728}. Transformer \cite{TRANSFORMER} is another burgeoning network block, which incorporates attention mechanism known for its ability to capture long-range dependencies. Built upon this, foundation models designed with promptable segmentation task emerged, such as segment anything model (SAM) \cite{SAM}. These models exhibit surprising versatility in error tolerant tasks. Furthermore, Ma et al. \cite{MEDSAM} transferred SAM to medical image scenario by fine-tuning the model on a large medical image dataset, and aimed to establish a universal model for medical image segmentation. Deep learning based segmentation can be expected to help physicians reduce time and may possess the potential for clinical applications. Despite remarkable achievements above, deep learning based algorithms, when in shortage of sufficient, various data and consistent labels of high quality, are by far still facing challenges in out-of-distribution generality, such as in cross-center and cross-anatomy applications.

\begin{figure}[htbp]
    \centering
    \includegraphics[width=\textwidth]{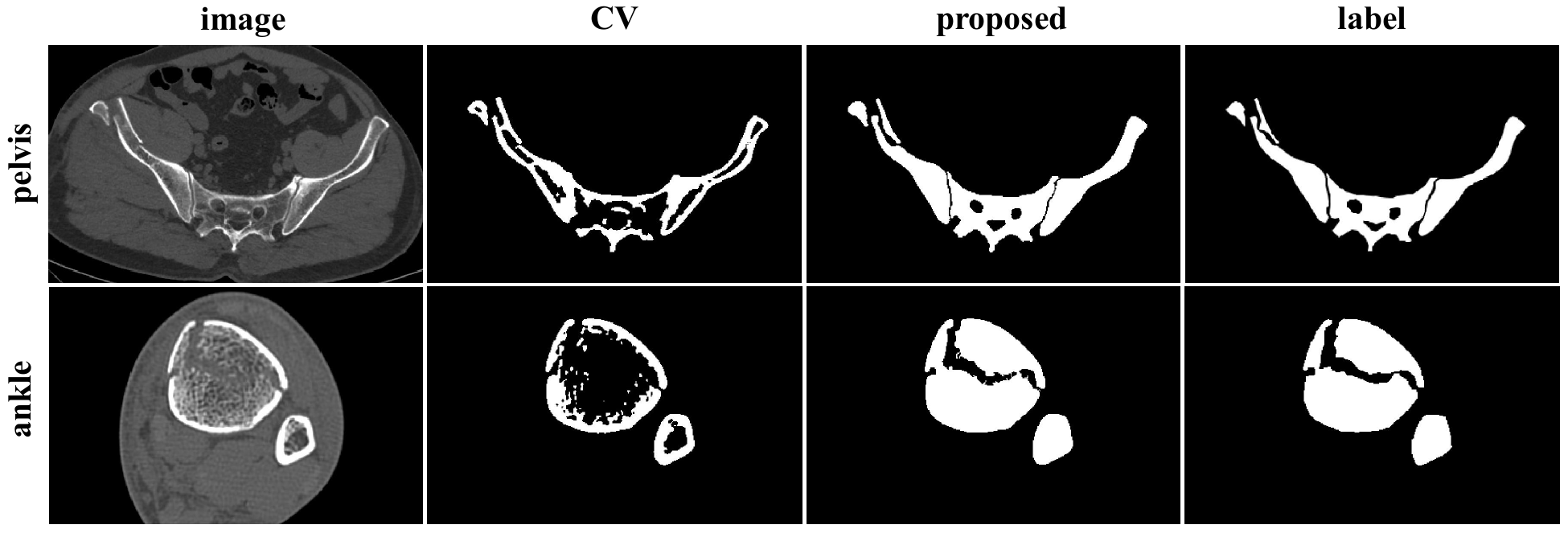}
        \vspace{-3mm}
    \caption{The performance of CV model and the proposed model when applied to bone CT images, such as the pelvis and the ankle.}
    \label{fig1}
\end{figure}

In contrast, traditional model based algorithms are distinguished for their advantages in mathematical interpretability and minimal data requirements. One such method is the active contour model \cite{MATHEMATICAL_PROBLEMS_IN_IMAGE_PROCESSING}, widely used in many segmentation tasks. The key idea is to achieve a contour that minimizes the designed energy functional, in which prior knowledge and information about the target can be easily embedded. The model can be formulated in terms of level set function or characteristic function, and be solved by optimization methods. According to the energy functional, active contour models can be broadly categorized into two types: region-based models \cite{CHAN_VESE,CONTINUOUS_MAX_FLOW_AND_MIN_CUT,PIECEWISE_POLYNOMIAL,ICTM,RSF,YANG2021107985,HAN2020107520,MIN201969} and edge-based models \cite{GAC,FAC,Cohen1997,Chen2017,Duits2018,Chen2024,SHAN2024110007}. For instance, some region-based models, such as Chan--Vese (CV) model \cite{CHAN_VESE} and continuous max-flow (CMF) model \cite{CONTINUOUS_MAX_FLOW_AND_MIN_CUT}, approximate the image foreground and background with two appropriate constants. Based on the CV model, in \cite{PIECEWISE_POLYNOMIAL}, the idea was proposed to approximate the image with piecewise polynomials, which can be seen as a simple framework of piecewise-polynomial Mumford-Shah model for image segmentation. The authors of \cite{ICTM} proposed an iterative convolution-thresholding method, where the contour length term is approximated by heat kernel convolution. Due to the piecewise constant approximation, CV and CMF models are more suitable to segment homogeneous objects. To handle intensity inhomogeneity, a region-scalable fitting (RSF) energy functional was designed in \cite{RSF}. Furthermore, the RSF term was combined with a transcendental constraint term in \cite{YANG2021107985}, which can supervise the level set function with a mutual segmentation atlas. The authors of \cite{HAN2020107520} proposed a new data fitting term based on Jeffreys divergence instead of widely used Euclidean distance and achieved relatively better performance. However, bone intensity cannot be uniformly characterized. Bone tissue consists of two types: trabecular bones and cortical bones. Trabecular bones are usually physiologically low in density and share similar intensity and texture with the surrounding tissues. On the other hand, cortical bones are denser and appear brighter \cite{IMAGING_OF_BONES_AND_JOINTS}. The differences of trabecular and cortical bones are intrinsic and therefore lead these models to segment only cortical bones, resulting in incomplete bone masks, as shown in figure \ref{fig1}.

For edge-based models, Caselles et al. \cite{GAC} proposed the classical geodesic active contour (GAC) model, which finds the optimal contour that minimizes its length weighted by an edge-detector function. While effective, this model was later extended to anisotropic metrics by Finsler active contours \cite{FAC}, enabling more flexible direction-dependent pattern detection. Beyond local minimization, the minimal path approach \cite{Cohen1997} applied a static Hamilton-Jacobian-Bellman PDE framework to find the global minimum of the GAC model, inspiring a series of models such as Finsler elastica \cite{Chen2017}, Reeds-Shepp car model \cite{Duits2018} and region-based Randers geodesic model \cite{Chen2024}.  The author of \cite{SHAN2024110007} proposed a compressively sensed geodesic active contour model, ensuring simultaneous sparse edge detection and accurate segmentation.

Meanwhile, some researchers also proposed specialized active contour algorithms. GAC model is a good candidate for bone segmentation due to its edge-driven nature. The authors of \cite{1663773} presented a geodesic active contour model constrained by global shape priors. In \cite{4906670}, the authors incorporated wavelet processing, automated seed growing and active contour models for pelvic CT segmentation. An active contour framework was proposed for bone micro-CT segmentation in \cite{KORFIATIS2017358}. A two-stage method proposed in \cite{GANGWAR201895} was applied to eliminate the spurious bridges cross thin cartilage interfaces. However, few articles fully take the influence of bone fractures and soft tissue edges into account. 

In this work, our aim is to develop a new geodesic active contour algorithm tailored for bone segmentation to fill this gap. To this end, we revisit the classical GAC model and address the following problems that inhibit its performance when applied to bone CT images:
\begin{itemize}
    \item[(i)] Edge obstruction. The contour evolution  might be obstructed by the edges of other organs or tissues, such as the intestines or bladder in pelvic images, potentially leading to undesired results.
    \item[(ii)] Edge leakage. The evolution speed is not easy to control. Under certain parameters, the contour might pass through bone edges and leak into the bones.
    \item[(iii)] Bone fracture, if presents, could provide no edges for the contour to stop.
\end{itemize}

Motivated by these, we propose a fracture interactive geodesic active contour (FI-GAC) algorithm integrating orthopedic domain knowledge and distance information to address these problems. Specifically, we observe that the bone and the soft tissue are intrinsically different in density and assume that this distinction could be reflected and quantified in image intensity, through the orthopedic domain knowledge of their CT values and the gray-scale mapping. Hence, we construct a novel edge-detector function that could help the algorithm to discriminate the bone and the soft tissue. Moreover, we find that the distance to the bone edges can be naturally incorporated as an adaptive step size to pace and stabilize the contour evolution. To handle the bone fracture, we further embed the fracture prompts into the distance information so that the contour can stop on the fractures. Main contributions of this work are summarized as follows: 
\begin{itemize}
    \item[(i)] To the best of our knowledge, this is the first literature that proposes a fracture interactive geodesic active contours for bone segmentation, which is intrinsically robust to bone fractures and soft tissue edges without depending on the annotated data. We apply the proposed algorithm in both pelvic and ankle segmentation to demonstrate its performance.
    \item[(ii)] We construct a novel edge-detector function that incorporates orthopedic domain knowledge to discriminate bone and soft tissue edges. It guides the contour towards the bones without being influenced by other soft tissues and considerably reduces mis-segmentation.
    \item[(iii)] We introduce distance information into the contour evolution as an adaptive step size to help the evolution slow down and stop, which stabilizes the evolution.
    \item[(iv)] We design a kind of fracture interaction that embeds manual prompts into the distance information to enable the contour to stop on the fractures, improving the accuracy with respect to the fractures.
\end{itemize}

The rest of this paper is organized as follows. In section \ref{Section 2}, we briefly revisit the classical geodesic active contour model and analyze several problems that inhibit its performance in bone segmentation. In section \ref{Section 3}, we detail the proposed FI-GAC algorithm integrating orthopedic domain knowledge and distance information. In section \ref{Section 4}, we demonstrate the performance of the proposed algorithm on both pelvic and ankle images, and provide a referable principle of easy parameter tuning. Finally, we conclude this work in section \ref{Section 5}.

\section{Preliminary}
\label{Section 2}

To begin with, the classical GAC model \cite{GAC,GAC_REVIEW} is revisited. Let $\Omega\subset\mathbb R^2$ be a bounded connected open subset, and $I:\Omega\to\mathbb R$ be a given gray-scale image. Let $\mathcal{C}$ be the set of piecewise continuously differentiable closed contours in $\Omega$, defined by
\[
    \mathcal{C} := \bigl\{c:[a,b]\to\Omega\ |c\text{ is piecewise continuously differentiable and } c(a)=c(b)\bigr\}.
\]
The GAC model is to solve the following minimization problem in the set $\mathcal{C}$,
\begin{equation}\label{J1}
    \min_{c\in\mathcal{C}}J_1(c):=\int_a^b g(|\nabla I(c(\theta))|)|c'(\theta)|\mathrm{d}\theta,
\end{equation}
where the energy $J_1(c)$ can be interpreted as the weighted length of contour $c$. Here, the function $g:[0,+\infty)\to(0,+\infty)$ is monotonically decreasing and vanishing at the positive infinity, usually taken as $g(z)=1/(1+z^2)$, in this situation
\[g(|\nabla I|) = \frac{1}{1+|\nabla I|^2}.\]More generally, it can be expressed in the form $g(z)=1/(1+f(z))$, where $f(0)=0$ and $f(z)\to +\infty$ as $z\to+\infty$. It can be observed that the weight $g(|\nabla I(x)|)$ takes small values when $x\in\Omega$ belongs to edges with large gradient norm. Therefore, when minimizing the energy functional $J_1$, the contour is forced to edges with large gradient norm. The weight $g(|\nabla I|)$ encodes image edge information and is named edge-detector function. Before the gradient operator, Gaussian smoothing can be optionally applied to reduce image noise, but this could also eliminate slight features.

Usually, the energy functional $J_1$ can be reformulated in terms of level set function $\phi:\Omega\to\mathbb R$, which represents the contour $c$ implicitly by its zero-level set $\{x\in\Omega\ |\phi(x)=0\}$. In the following discussion, we adopt the convention that $\phi$ is negative inside the contour $c$ and positive outside. A main advantage of this representation is that topological changes of the contour are automatically taken into account. The minimization problem is rewritten as
\begin{equation}\label{J2}
    \min_{\phi}J_2(\phi):=\int_{\Omega}g(|\nabla I(x)|)|\nabla H(\phi(x))|\mathrm{d}x,
\end{equation}
which can be proven equivalent to \eqref{J1} by co-area formula. Here, the $H$ denotes the Heaviside function by  
\begin{equation*}
H(z) := \left\{
    \begin{aligned}
        1, && z\in[0,+\infty),\\
        0, && z\in(-\infty,0).
    \end{aligned}\right.
\end{equation*}

In particular, an area term can be added onto \eqref{J2}, which is defined by
\[A(\phi):=\alpha\int_{\Omega}g(|\nabla I(x)|)H(-\phi(x))\mathrm{d}x,\]
where the parameter $\alpha$ is a balancing weight. This term contributes to speeding up the contour evolution and handling situations with non-convex targets. To be clear, when $\alpha$ is taken positive, minimizing $A(\phi)$ means minimizing the weighted area inside the contour $c$ and helps the contour to shrink, vice versa when $\alpha$ is negative. It is noteworthy that if $\alpha$ is taken too large, the evolution speed is not easy to control and the contour might easily pass through the target edges, since the area term plays a major role during the minimization. Therefore, the parameter $\alpha$ should be tuned very carefully.

To solve problem \eqref{J2} numerically, one can consider a smoothed version of the problem, written as
\begin{equation}\label{SMOOTHED_VERSION}
    \min_{\phi}J_{2,\epsilon}(\phi)+A_\epsilon(\phi),
\end{equation}
where $J_{2,\epsilon}(\phi)$ is actually $J_2(\phi)$ that replaces the Heaviside function $H$ with a smooth approximation $H_\epsilon$, and the same for $A_\epsilon(\phi)$. The corresponding derivative of $H_\epsilon$ denotes $\delta_\epsilon$. One possible option for $H_\epsilon$ and $\delta_\epsilon$ is
\[
H_\epsilon(z) = \frac12\left(1+\frac{2}{\pi}\arctan\left(\frac{z}{\epsilon}\right)\right) \quad\text{and}\quad \delta_\epsilon(z) = \frac{1}{\pi}\frac{\epsilon}{\epsilon^2+z^2}, 
\]
respectively. By introducing an artificial time $t\ge 0$ to the level set function $\phi$, one can deduce the associated gradient flow equation of \eqref{SMOOTHED_VERSION}, which is
\begin{equation}\label{GRADIENT_FLOW}
\left\{
    \begin{aligned}
        &\frac{\partial\phi}{\partial t} = \left\{\mathrm{div}\left(g\frac{\nabla\phi}{|\nabla\phi|}\right)+\alpha g\right\}\delta_\epsilon(\phi),&&\text{in}\ (0,+\infty)\times\Omega,\\
        &\phi(0,x) = \phi_0(x),&&\text{in}\ \Omega,
    \end{aligned}\right.
\end{equation}
where $\mathrm{div}$ means the divergence operator.
Hereafter, we denote $g(|\nabla I|)$ as $g$ for simplicity when the context is clear. In the above PDE, the factor $\delta_\epsilon(\phi)$ can be replaced alternatively by $|\nabla\phi|$. This would affect only the speed of gradient descent but not the direction \cite{A_VARIATIONAL_LEVEL_SET_APPROACH}, which is also adopted in this work. The function $\phi_0$ is an initial level set function with respect to the initial contour $c_0$. A good candidate of $\phi_0$ is the signed distance function defined as
\begin{equation*}
\phi_0(x) = \left\{
    \begin{aligned}
        &+d(x, c_0), && \text{if $x$ lies outside $c_0$},\\
        &-d(x, c_0), && \text{if $x$ lies inside $c_0$},
    \end{aligned}\right.
\end{equation*}
where $d(x, c_0)$ is the Euclidean distance from $x$ to $c_0$. During the evolution, the level set function needs to be periodically reinitialized to a signed distance function, so as to maintain good computational performance. DRLSE \cite{DRLSE} proposed a regularization energy functional which enables the level set function to remain a signed distance function automatically during the evolution. Finally, the flow equation \eqref{GRADIENT_FLOW} is discretized and solved in an iterative process. By a direct discretization, the classical GAC algorithm writes as the following algorithm \ref{CLASSICAL_GAC}.

\begin{algorithm}
    \caption{Classical GAC Algorithm}
    \label{CLASSICAL_GAC}
    \begin{algorithmic}[1]
    \Require Gray-scale image $I$, initial level set function $\phi_0$, parameter $\alpha$, time step size $h$, and iteration number $N$.
    \State Compute the edge-detector function $g \gets 1/(1+|\nabla I|^2)$.
    \For {$n := 0\ \mathrm{to}\ N-1$}
        \State $F_n \gets \mathrm{div}\left(g\nabla\phi_n/|\nabla\phi_n|\right)+\alpha g$, 
        \State $\phi_{n+1} \gets \phi_{n} + hF_n|\nabla\phi_n|$. 
    \EndFor
    \Ensure Final level set function $\phi_N$.
    \end{algorithmic}
\end{algorithm}

In bone segmentation, due to the physiological features of cortical and trabecular bones, GAC algorithm is more suitable because of its ability to capture edge information. We test the performance of algorithm \ref{CLASSICAL_GAC} in bone segmentation, and observe several problems that inhibit the potential of GAC algorithm. 
\begin{itemize}
    \item[(i)] Edge obstruction. In the classical GAC algorithm, edge-detector function will encode all edges in the image, including the edges of other organs or tissues. During the evolution, the contour is not capable to distinguish soft tissue edges from bone edges. Therefore, the contour is very likely to be obstructed.
    \item[(ii)] Edge Leakage. To deal with the non-convexity of bones, the area term is added to help the contour shrink. Nevertheless, if the coefficient is taken too large, this term will dominate the minimization and force the contour to keep shrinking even though it has reached the desired edge, causing the contour to pass through the bone edges and leak into the bone. This is because in this situation the area term dominates the energy functional and the vanishing contour is a local minimum of the area term.
    \item[(iii)] Bone fracture. Edge-detector function leads the contour into a local minimum where the edge is of large gradient norm. However, if bone fractures exist, there are possibly no obvious edges on the fractures so that the contour cannot stop.
\end{itemize}
   
\begin{figure}[htbp]
    \centering
    \includegraphics[width=\textwidth]{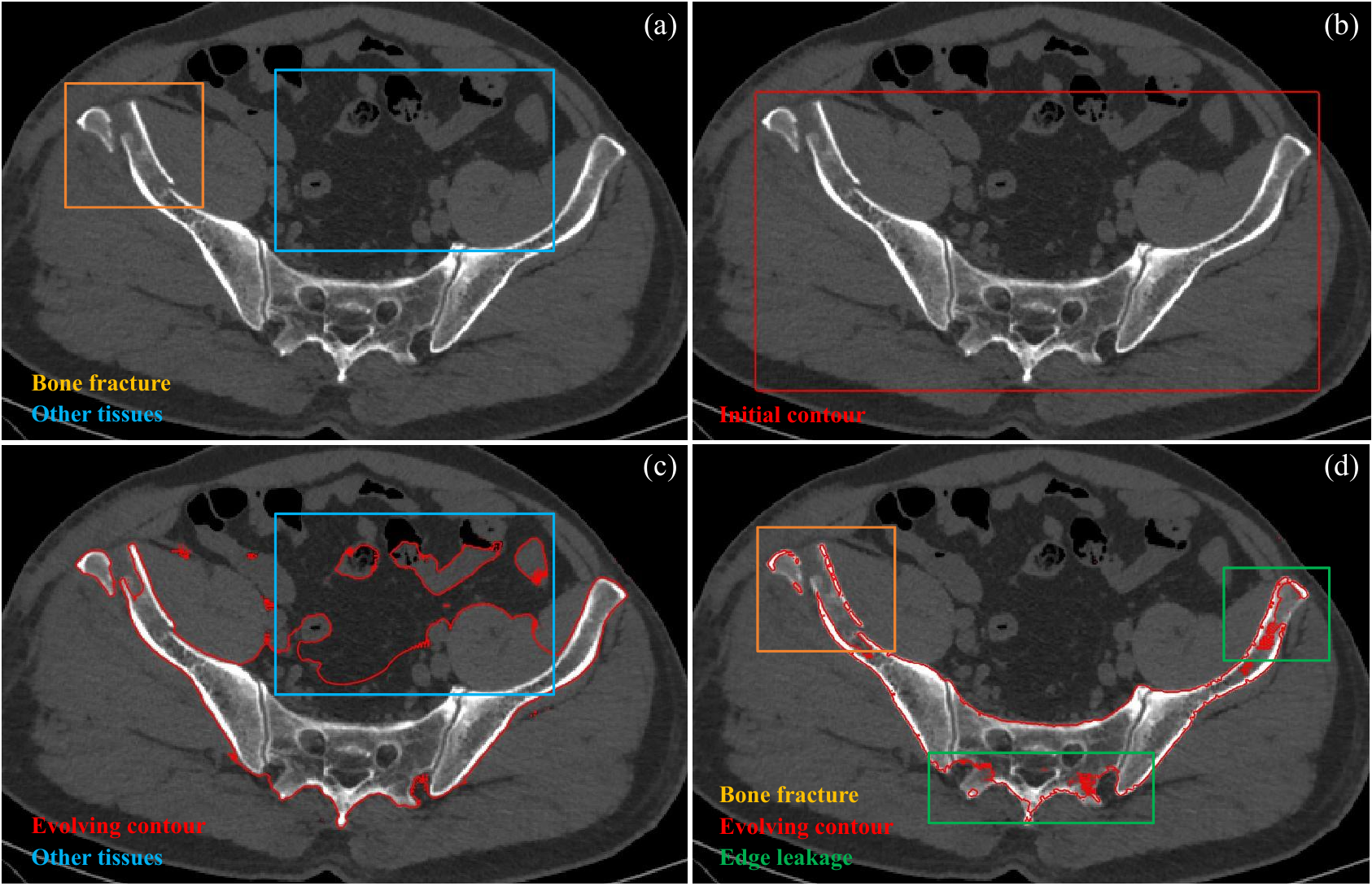} 
        \vspace{-3mm}
    \caption{An illustration of several problems of the classical GAC algorithm in pelvic segmentation. All red lines represents contours during the iterations. Figure 2(a) is a pelvic CT image with the existence of bone fracture and other tissues, marked by the orange box and the blue box respectively. Figure 2(b) shows the initial contour. Figure 2(c) illustrates that the evolving contour is obstructed by other tissues. Figure 2(d) demonstrates that: first, as shown in the orange box, the evolving contour does not stop on the fractures since no obvious edges exist; second, as shown in the green boxes, the contour passes through the edges and leaks into the bone in very late iterations. }
    \label{fig2}
\end{figure}

 Figure \ref{fig2} is an illustration of these problems in a pelvic CT image. These problems are the starting points and guidelines in the design of our algorithm.

\section{Methods}
\label{Section 3}

In this section, we detail the proposed FI-GAC algorithm whose pipeline is shown in figure \ref{fig4}.

\subsection{Integrating domain knowledge}
For raw CT images, each pixel is the CT value that represents the relative X-ray attenuation coefficient of a substance compared to that of water, i.e.,
\[\mathrm{CT\ value}=\frac{\mu_{\mathrm{substance}}-\mu_{\mathrm{water}}}{\mu_{\mathrm{water}}}\times 1000,\]
in Hounsfield unit (HU). Water takes a CT value of $0$ HU, while air typically takes $-1000$ HU. When displaying the CT images in gray-scale, radiologists need to choose an appropriate CT window to truncate the whole spectrum and remap the value into the range of $[0,255]$. The window width and window level are denoted by $W_w$ and $W_l$ respectively, then the gray-scale mapping can be expressed by
\begin{equation}
    m\left(z;W_w,W_l\right)=\left\{
    \begin{aligned}
        &255, && z\in\left[W_l+\frac{W_w}{2},+\infty\right),\\
        &\frac{255}{W_w}\left(z-W_l\right)+\frac{255}{2}, && z\in\left[W_l-\frac{W_w}{2}, W_l+\frac{W_w}{2}\right),\\
        &0, && z\in\left(-\infty, W_l-\frac{W_w}{2}\right).
    \end{aligned}\right.\nonumber
\end{equation}
Usually, the adjustment of CT window depends on various factors, such as the clinical problem, the concerned anatomical region, and the desired diagnostic information.

In this work, bone segmentation is considered under bone window, which corresponds to orthopedic diagnosis. Generally speaking, there is not a rigid bone window that can be applied to all situations. But fortunately, the recognized technical specifications for CT examination do exist, and for a specific anatomy, there is a broad range of practicable CT windows \cite{CT_WINDOWS2,CT_WINDOWS1}, which is adequate for the proposed algorithm. It is noticed that bones and surrounding soft tissues are intrinsically different in density and there should be a gap between their CT values. Therefore, the gap between their gray-scale values can be naturally derived if bone window is given. This orthopedic domain knowledge can be integrated into the GAC algorithm to help distinguish bones from soft tissues.

It is known that the practicable range for bone window width and level is $W_w\in[w_1,w_2]$ and $W_l\in[l_1,l_2]$, where the unit HU is omitted. Suppose that the CT value has a lower bound $B$ for bones, and has an upper bound $S$ for soft tissues. Compute the corresponding gray-scale value bounds by the following two optimization problems
\begin{equation}\label{theta1}
    \begin{aligned}
        \theta_1 = &\max~m\left(z;W_w,W_l\right)\\
        &~\mathrm{s.t.}~z\le S,w_1\le W_w\le w_2,l_1\le W_l\le l_2,
    \end{aligned}
\end{equation}
\begin{equation}\label{theta2}
    \begin{aligned}
        \theta_2 = &\min~m\left(z;W_w,W_l\right)\\
        &~\mathrm{s.t.}~z\ge B,w_1\le W_w\le w_2,l_1\le W_l\le l_2.
    \end{aligned}
\end{equation}

\begin{proposition}\label{Proposition}
    Suppose $S\le l_1$ and $w_1>0$. If one of the following inequalities \[B\ge l_2\quad\mathrm{or}\quad\frac{S-l_1}{w_2}\le\frac{B-l_2}{w_1}\] holds, then $\theta_1\le\theta_2$.
\end{proposition}\label{prop}
\begin{proof}
    The assumptions for $S\le l_1$ and $w_1>0$ are natural, since soft tissues usually have lower CT values than bones and $w_1>0$ is the lower bound of window width. Because $S\le l_1$, problem \eqref{theta1} attains its maximum
    \[\theta_1=m(S;l_1,w_2)=\frac{255}{w_2}\left(S-l_1\right)+\frac{255}{2}.\]
    Next, consider the following two situations: 
    \begin{itemize}
        \item If $B\ge l_2$, problem \eqref{theta2} attains its minimum
        \[\theta_2=m(B;l_2,w_2)=\frac{255}{w_2}\left(B-l_2\right)+\frac{255}{2},\]
    and it is easy to obtain $\theta_1\le\theta_2$.
        \item If $B<l_2$, problem \eqref{theta2} attains its minimum
    \[\theta_2=m(B;l_2,w_1)=\frac{255}{w_1}\left(B-l_2\right)+\frac{255}{2},\]
    then $\theta_1\le\theta_2$ still holds because of the second inequality.
    \end{itemize}
\end{proof}

To better capture bone features, we propose to integrate both the bone intensity and gradient norm information into the edge-detector function, generally expressed in the form
\begin{equation}\label{EDGE-DETECTOR_general}
    \Tilde{g} = \Tilde{g}(I, |\nabla I|). 
\end{equation}
A specific example can be taken as 
\begin{equation}\label{EDGE-DETECTOR}
    \Tilde{g}(I, |\nabla I|) = \frac{1}{1+f_1(I)}+\frac{\gamma}{1+f_2(|\nabla I|)},
\end{equation}
where $f_1,f_2$ are two high-pass filters and $\gamma>0$ is an adjustable weight to balance two components. In this work, pick any thresholds $\varepsilon_1\in[\theta_1,\theta_2]$, $\varepsilon_2\in[0,\theta_2-\theta_1]$ and choose
\[f_1(z)=\bigl((z-\varepsilon_1)_+\bigr)^{\delta_1},\quad f_2(z)=\bigl((z-\varepsilon_2)_+\bigr)^{\delta_2},\] where $z_+ = \max\{z,0\}$ and $\delta_1,\delta_2 > 0$. Practically, the conditions in Proposition 1 are easily checked and satisfied, which ensures the existence of $\varepsilon_1$ and $\varepsilon_2$.

\begin{figure}[htbp]
    \centering
    \includegraphics[width=\textwidth]{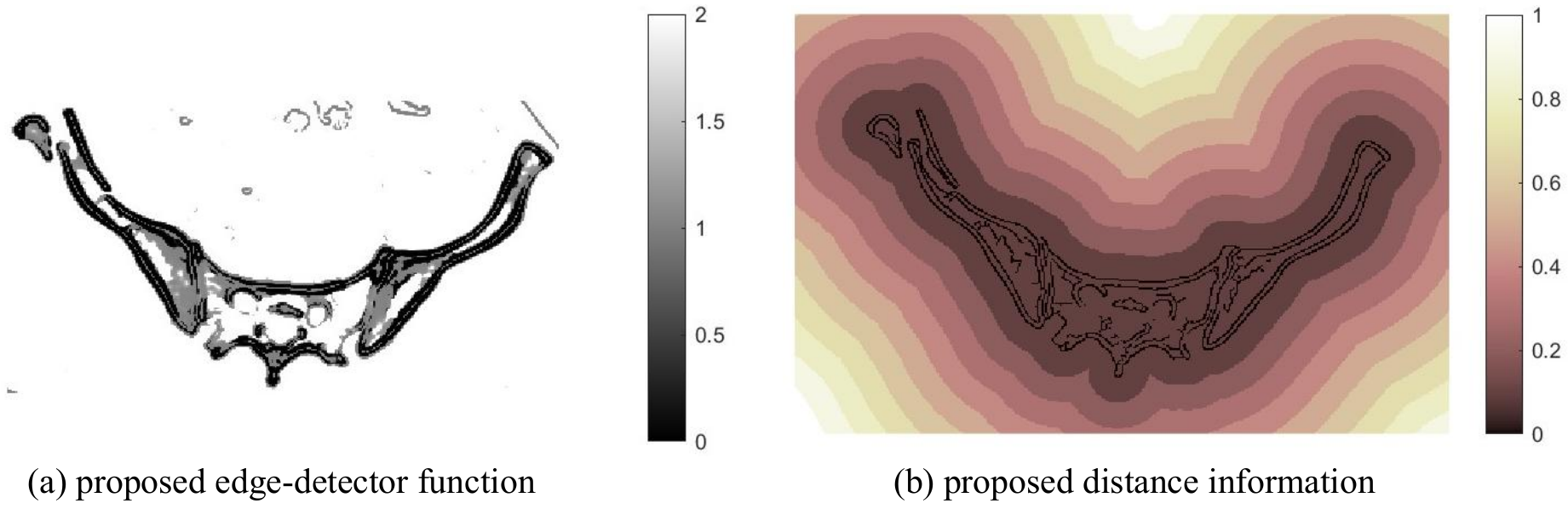}
    \vspace{-3mm}
    \caption{An illustration of : (a) the proposed edge-detector function; (b) the proposed distance information.}
    \label{fig3}
\end{figure}

Hence, we construct a novel edge-detector function, which takes both intensity and gradient norm into consideration, and filters out soft tissues and very weak edges. The proposed edge-detector function can guide the contour to bone edges with high intensity and large gradient norm without influenced by other soft tissues. Figure \ref{fig3}(a) is an illustration of the proposed edge-detector function, in which the information about the target is well extracted and highlighted.

\subsection{Integrating distance information and fracture prompts}
As mentioned in section \ref{Section 2}, a large value of $\alpha$ contributes to speeding up the contour evolution, but will easily cause edge leakage. This can be addressed by choosing a small value of $\alpha$ or updating $\alpha$ during the iteration, but both of these methods are not flexible enough. Inspired by edge extraction algorithms, such as Canny algorithm, if bone edges can be extracted, a pixel-wise map can be derived by computing the distance from each pixel to the bone edges. Then, this map is called distance information and can be used as an adaptive step size to pace the contour evolution. That is to say, when the contour is far from the bone edges, the evolution is fast, and vice versa. During the evolution, the contour gradually slows down and eventually stops at the bone edges.

Suppose that all edges have been obtained in the image by some edge extraction algorithm, and denote the set of all edge points by $E_{\mathrm{all}}$. By far, the set $E_{\mathrm{all}}$ contains all possible edges and only bone edges are needed. Note that in a small neighbourhood of bone edges the pixels usually share high intensity, which is not the case for non-bone edges. By this property, non-bone edges can be filtered out and bone edges are extracted. Denote the set of extracted bone edge points by $E_{\mathrm{bone}}$, then its characteristic function is expressed by
\begin{equation}\label{Ebone}
    \chi_{E_{\mathrm{bone}}} = \chi_{E_{\mathrm{all}}}\cdot\chi_{\{\phi_0\le0\}}\cdot\chi_{\{I*K\ge\eta\}}, 
\end{equation}
where $\{\phi_0\le 0\}$ represents the inside of the initial contour and can be regarded as the region of interest (ROI), $K$ is a convolution kernel and $\eta$ is a threshold to filter out non-bone edges. Thus, the proposed distance information factor can be written as  
\begin{equation*}
    \beta(x)=\dfrac{d(x,E_{\mathrm{bone}})}{\max_{y\in\Omega}d(y,E_{\mathrm{bone}})},\quad x\in\Omega, 
\end{equation*}
where by dividing the maximum, the distance is normalized into $[0,1]$. 

Bone fracture, if presents, could provide no obvious edges for the contour to stop. In this case, the classical GAC algorithm can hardly find the fractures. In the proposed algorithm, fracture prompts, i.e., manual annotations on the fractures, are embedded into the distance information and help the contour stop on the fractures. This provides a way for the surgeons to pay attention to and interact with bone fractures in GAC algorithm. Suppose the set of fracture prompts is denoted by $E_{\mathrm{prompt}}$. Update the set of bone edges as
\[\Tilde{E}_{\mathrm{bone}}=E_{\mathrm{bone}}\cup E_{\mathrm{prompt}},\]
then compute the corresponding distance information factor
\begin{equation}\label{DISTANCE-INFORMATION}
    \Tilde{\beta}(x) = \dfrac{d(x,\Tilde{E}_{\mathrm{bone}})}{\max_{y\in\Omega}d(y,\Tilde{E}_{\mathrm{bone}})},\quad x\in\Omega.
\end{equation}
Note that $\Tilde{\beta}(x) = 0$ for $x\in\Tilde{E}_{\mathrm{bone}}$, which stops the contour evolution. Figure \ref{fig3}(b) is an illustration of the proposed distance information, and the equidistant lines are drawn.

\subsection{Overview of the proposed algorithm}
In our algorithm, the energy functional is depicted as follows:
\begin{equation*}
\min_{\phi}~\int_{\Omega}\Tilde{g}(I(x), |\nabla I(x)|)|\nabla H(\phi(x))|\mathrm{d}x + \alpha\int_{\Omega}\Tilde{g}(I(x), |\nabla I(x)|)H(-\phi(x))\mathrm{d}x, 
\end{equation*}
where $\Tilde{g}$ is the proposed edge-detector function integrated with domain knowledge. Multiplied by the distance information $\Tilde{\beta}$ as an adaptive step size, the corresponding gradient flow equation turns to be
\begin{equation}\label{NEW_GRADIENT_FLOW}
\left\{
    \begin{aligned}
        &\frac{\partial\phi}{\partial t} = \Tilde{\beta}\left\{\mathrm{div}\left(\Tilde{g}\frac{\nabla\phi}{|\nabla\phi|}\right)+\alpha \Tilde{g}\right\}|\nabla\phi|,&&\text{in}\ (0,+\infty)\times\Omega,\\
        &\phi(0,x) = \phi_0(x),&&\text{in}\ \Omega.
    \end{aligned}\right.
\end{equation}
The above equation is then solved by discretization and iterations. The proposed edge-detector function $\Tilde{g}$ guides the contour towards bone edges without influenced by other organs or tissues, and the distance information factor $\Tilde{\beta}$ helps the contour slow down and stop at bone edges and fractures.

\begin{figure}[htbp]
    \centering
    \includegraphics[width=\textwidth]{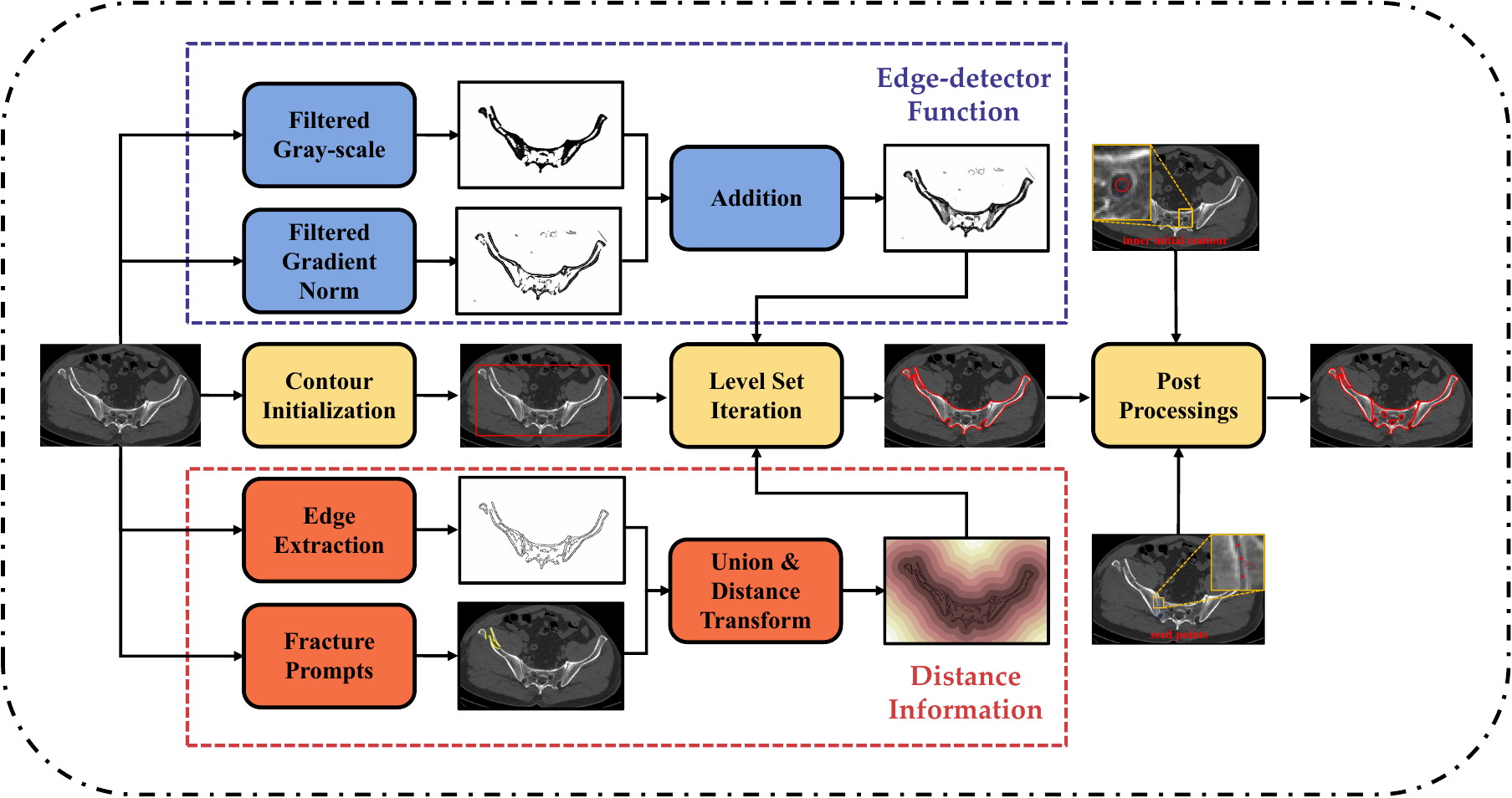} 
    \vspace{-3mm}
    \caption{The overall pipeline of the proposed FI-GAC algorithm integrating domain knowledge and distance information.}
    \label{fig4}
\end{figure}

As shown in figure \ref{fig4}, the overall pipeline of the proposed algorithm consists of the following steps:
\begin{itemize}
    \item[(i)] Contour initialization. Either manual or auto initialization can be applied. Auto
initialization can be realized by any pre-trained deep learning object detector represented by R-CNN and YOLO, which will remarkably reduce time spent on initialization. Alternatively, a simple way to determine the bounding box is finding the extreme points of the high intensity region. The inside of initial contour is served as the ROI, which is used to extract bone edges.
    \item[(ii)] Computation of the proposed edge-detector function. For the concerned bone anatomy, such as pelvis or ankle, determine the range of practicable bone windows and the range of CT values for the bone and the soft tissue. Calculate the thresholds $\varepsilon_1$, $\varepsilon_2$, and then compute the edge-detector function $\Tilde{g}$ by \eqref{EDGE-DETECTOR}.
    \item[(iii)] Computation of the proposed distance information. Extract all edges by any algorithm such as Canny algorithm, and filter out non-bone edges. Combined with fracture prompts, the distance information $\Tilde{\beta}$ is computed by \eqref{DISTANCE-INFORMATION}.
    \item[(iv)] Level set iteration. Discretize \eqref{NEW_GRADIENT_FLOW} and iterate $N$ steps.
    \item[(v)] Post processings. After iterations, a preliminary mask can be obtained. If there are inner holes or narrow gaps in the bone structure, post processings are needed because GAC contour is hard to get into the inner holes and narrow gaps due to its nature of length minimization. As shown in figure \ref{fig4}, inner initial contours are drawn to expand under a negative $\alpha$ so as to find the inner holes, and narrow gaps are found by selecting seed points in the narrow gaps and then applying region growing algorithm locally. 
\end{itemize}

\section{Experiments and results}
\label{Section 4}

In this section, experiments on both pelvic and ankle images are conducted to test the performance of the proposed algorithm. An elaboration is given on the implementation details and evaluation metrics used in the experiments. The results are presented in both visual and quantitative aspects, showing promising performance in pelvis and ankle segmentation. Principles for parameter tuning are provided in the discussion.

\subsection{Implementation details}
In the experiments, we test CT images from three sources: (i) the open-sourced CTPelvic1K-dataset6-CLINIC \cite{DEEP_LEARNING_TO_SEGMENT_PELVIC}; (ii) the pelvic CT dataset collected by the Fourth Medical Center of Chinese PLA General Hospital; (iii) the ankle CT dataset collected by People's Hospital Peking University. 

For the construction of the proposed edge-detector function, we take the practical range of bone window width $W_w\in[1000,1500]$ and window level $W_l\in[250,350]$ for both pelvis and ankle. The CT value of soft tissues is considered no more than 100 HU and that of bones no less than 300 HU. Then, the conditions in proposition \ref{Proposition} are fulfilled. Next, calculate $\theta_1 = 102$ and $\theta_2 = 115$ in problem \eqref{theta1} and \eqref{theta2} respectively, and the thresholds $\varepsilon_1,\varepsilon_2$ can be flexibly picked in the range $\varepsilon_1\in[102,115]$ and $\varepsilon_2\in[0,13]$. For example, we choose $\varepsilon_1=102$ and $\varepsilon_2=13$ for both pelvis and ankle in our implementation, maximizing bone preservation and weak edge removal. We fix $\delta_1 = 1$, $\delta_2 = 2$ and $\gamma = 1$ in \eqref{EDGE-DETECTOR} throughout the experiments.

For the construction of distance information factor, Canny algorithm is applied to extract all edges in the image, where the double thresholds are taken as $0.2$ and $0.8$. The convolution kernel $K$ is taken a $3\times 3$ average kernel to filter out non-bone edges, namely, 
\begin{equation*}
    K = \frac19
\begin{pmatrix}
1&1&1\\
1&1&1\\
1&1&1
\end{pmatrix},
\end{equation*}
and the corresponding threshold $\eta$ in \eqref{Ebone} is empirically taken as $\eta = 70$. The contour evolution is initialized with the bounding box of the bone region, which can be roughly determined by finding the extreme points of the high intensity region. 

\subsection{Evaluation metrics}
Four metrics are used to evaluate the segmentation performance. Denote $\mathrm{G}$ and $\mathrm{S}$ the groundtruth and the segmentation result respectively. In the following expressions, $|\cdot|$ means the Hausdorff measure.
\begin{itemize}
    \item Dice and Jaccard coefficients. Evaluate the overlapping ratio between the groundtruth and the segmentation result, and describe the accuracy of pixel-wise prediction. Both are ranged in $[0\%,100\%]$, and higher coefficient indicates better segmentation. They are defined as 
    \[
    \mathrm{Dice}=2\frac{|\mathrm{G}\cap\mathrm{S}|}{|\mathrm{G}|+|\mathrm{S}|},\quad \mathrm{Jaccard}=\frac{|\mathrm{G}\cap\mathrm{S}|}{|\mathrm{G}\cup\mathrm{S}|}.
    \]
    \item Hausdorff distance (HD) and average symmetric surface difference (ASSD). Evaluate the difference between the boundary of groundtruth and the boundary of segmentation result. The lower these metrics are, the better the segmentation is. These two metrics are defined as 
    \begin{equation}
        \begin{aligned}
        &\mathrm{HD} = \max\left\{\max_{x\in\partial\mathrm{G}}d(x,\partial\mathrm{S}),\max_{y\in\partial\mathrm{S}}d(\partial\mathrm{G},y)\right\},\\
        &\mathrm{ASSD} = \frac{1}{|\partial\mathrm{G}|+|\partial\mathrm{S}|}\left\{\sum_{x\in\partial\mathrm{G}}d(x,\partial\mathrm{S})+\sum_{y\in\partial\mathrm{S}}d(\partial\mathrm{G},y)\right\}.
        \end{aligned}\nonumber
    \end{equation}
    Here the $\partial\mathrm{G}$ and $\partial\mathrm{S}$ denote the boundaries of $\mathrm{G}$ and $\mathrm{S}$, respectively. 
  
\end{itemize}

\subsection{Ablation study}
\label{ablation}
\begin{figure}[htbp]
    \centering
    \includegraphics[width=\textwidth]{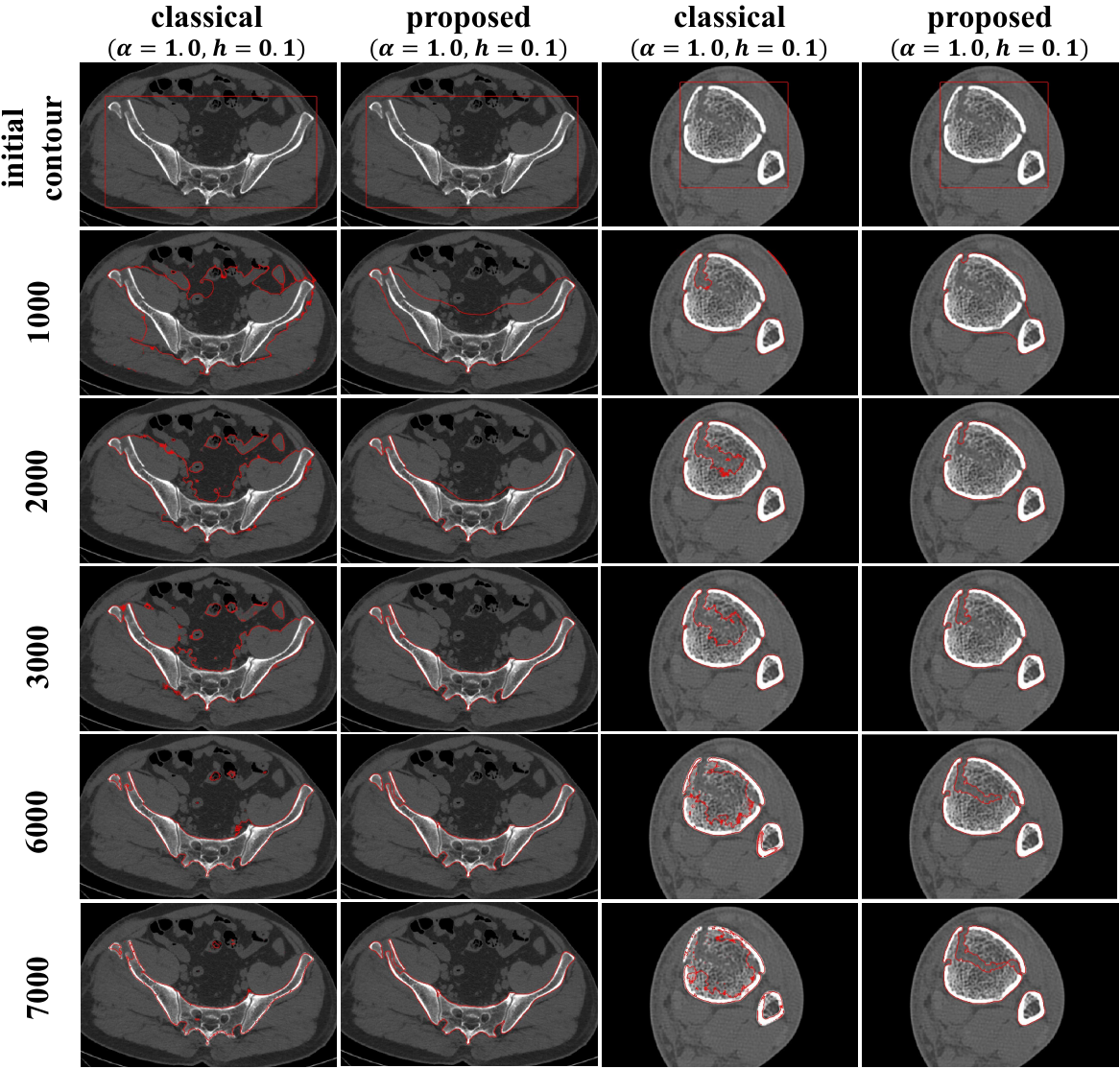}
        \vspace{-3mm}
    \caption{The behaviors of the proposed algorithm compared with the classical GAC algorithm under the same initial contour. For all of the four columns, the parameters are set identically as $\alpha=1.0$ and $h=0.1$.}
    \label{fig5}
\end{figure}

Above all, we are due to demonstrate that the behaviors of the proposed algorithm accord with our expectations, that is to say, the contour evolution overcomes the problems of edge obstruction, edge leakage and bone fracture. We apply our algorithm to both fractured pelvic and ankle images, and compare the behaviors of the evolving contour with the classical GAC algorithm under identical initial contour and parameters $\alpha=1.0,h=0.1$. The intermediate contours at 1000, 2000, 3000, 6000 and 7000 iterations are shown in figure \ref{fig5}. 
It can be observed that: first, for the proposed algorithm, the evolving contour is not obstructed by other organs and therefore mis-segmentation is avoided; second, in contrast to the classical GAC algorithm, where the contour is over-evolved and collapsed in very late iterations, the proposed algorithm enables the contour to stop at the bone edges and fractures, and maintains the shape for a relatively long stage of iterations without edge leakage, which makes the evolution stable and provides chances to stop the iteration. These results verify that the proposed algorithm is effective to solve the aforementioned problems.

\begin{figure}[htbp]
    \centering
    \includegraphics[width=\textwidth]{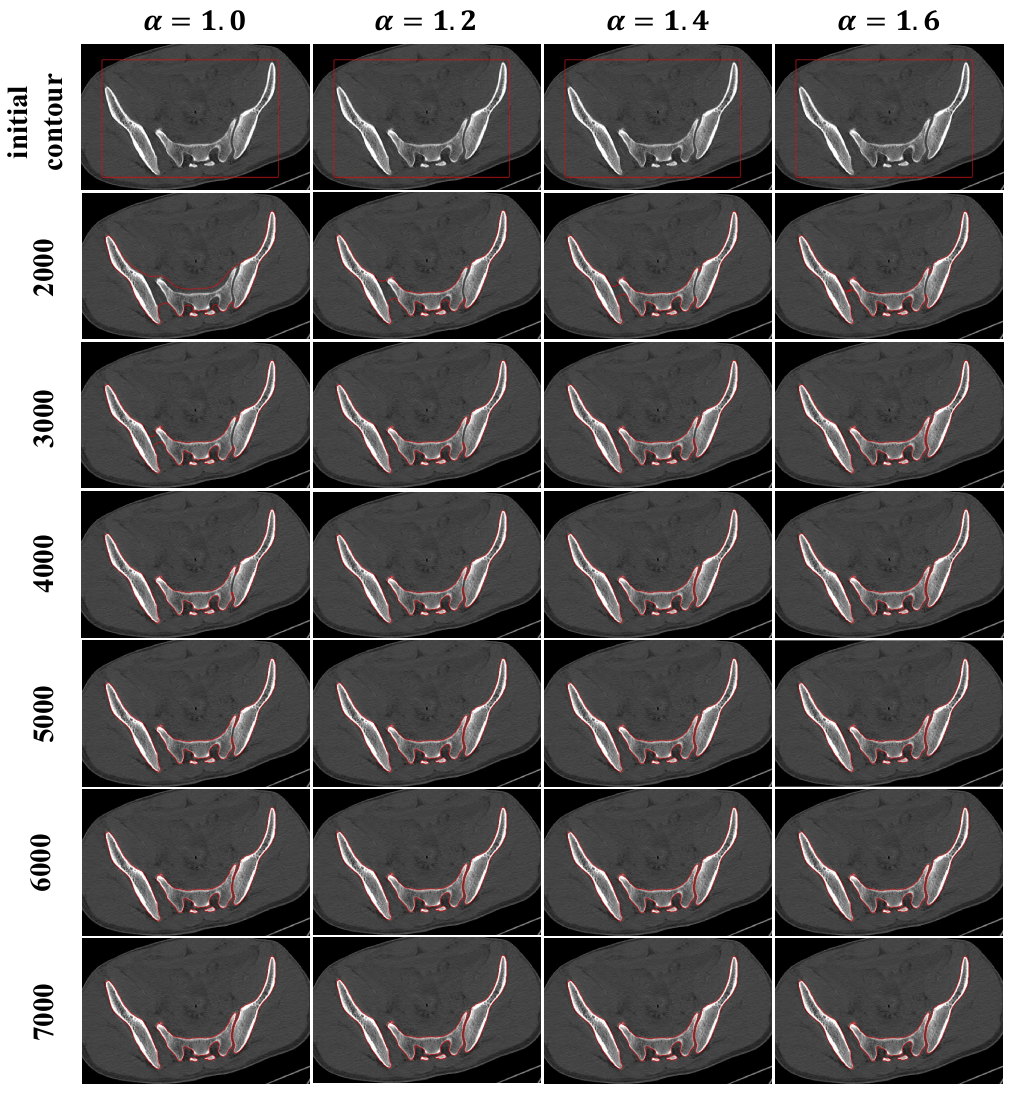}
    \vspace{-3mm}
    \caption{The robustness of the proposed algorithm with respect to $\alpha$ under identical $h=0.1$.} 
    \label{fig6}
\end{figure}

Next, as mentioned in the section \ref{Section 2}, the classical GAC algorithm is very sensitive to the choice of $\alpha$, since a large $\alpha$ will cause edge leakage. However, the proposed algorithm overcomes this problem and behaves robustly with respect to $\alpha$. We show the iterations of the proposed algorithm under $\alpha = 1.0,1.2,1.4,1.6$ and $h=0.1$, where the contour achieves the target at the 
$6300,4700,3800,3100$ iteration respectively and maintains the shape in the following iterations, as illustrated in figure \ref{fig6}, which sheds light on the robustness to $\alpha$. It can also be observed that a larger $\alpha$ contributes to speeding up the evolution.

\subsection{Comparisons with other methods}
\begin{figure}[htbp]
    \centering
   \includegraphics[width=\textwidth]{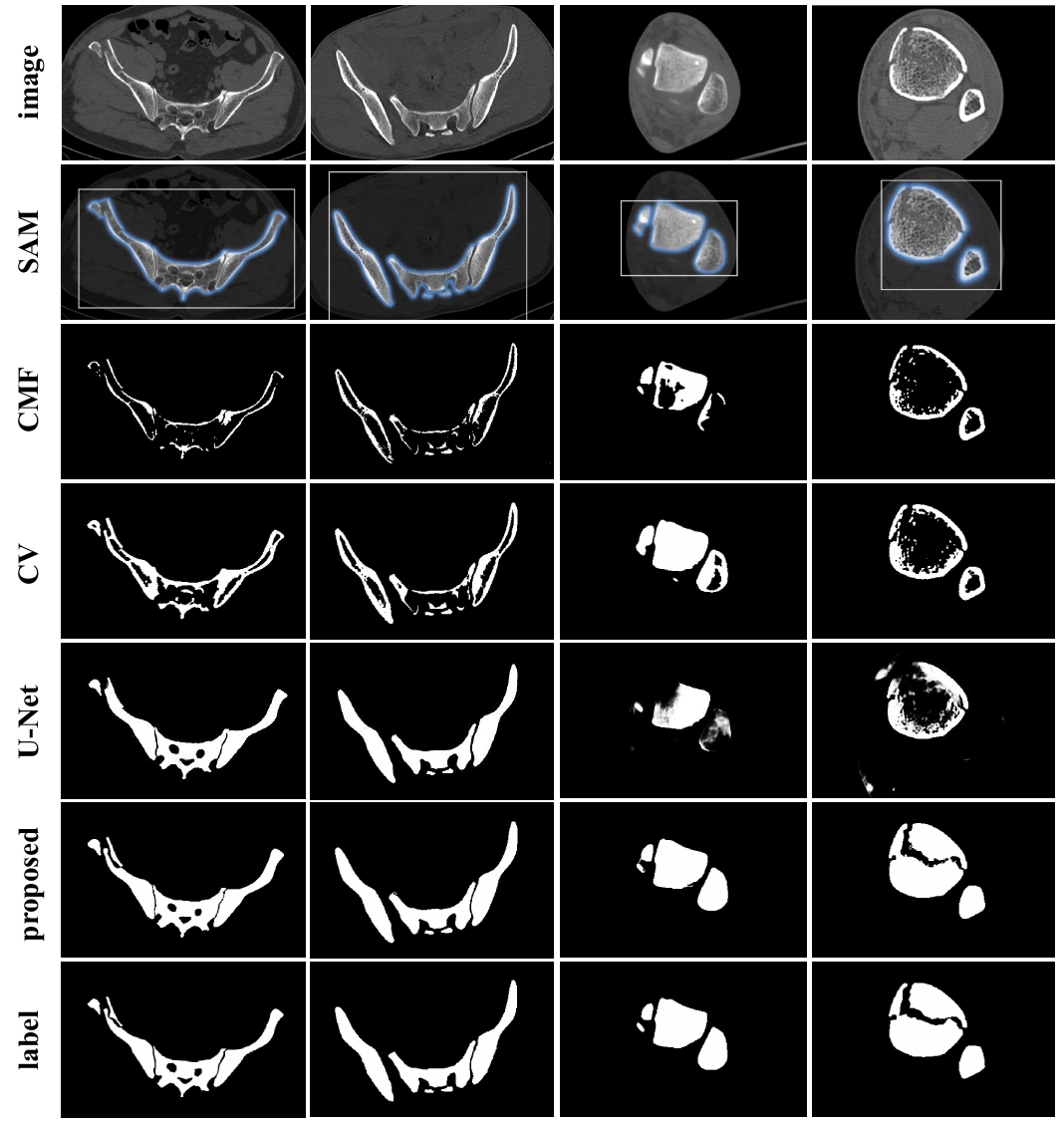}
   \vspace{-3mm}
    \caption{Visual comparisons of the segmentation masks between different methods. The second row is the screen-captured results of SAM, where the white boxes are the prompts and the blue contours represent the segmentation boundaries. For other methods, segmentation and groundtruth labels are shown.}
    \label{fig7}
\end{figure}

Additionally, we compare the segmentation masks between the proposed algorithm and other methods, such as SAM, U-Net, CMF and CV. The results of SAM are given by its demo on the official website\footnote{https://segment-anything.com} with box prompts and are then screen-captured. The usage of U-Net follows the original architecture \cite{UNET} which downsamples 4 times with intermediate channels $[64,128,256,512,512]$, and is trained on a NVIDIA A6000 GPU (24GB) with BCE loss and Adam optimizer (learning rate $=1\times10^{-3}$, batch size $=8$) for 100 epochs. Training data is a subset of the dataset (i) (90 volumes, around 30000 slices). The results of CMF and CV models shown in figure \ref{fig7} are the best results among multiple feasible parameters. In figure \ref{fig7}, we can observe that:
\begin{itemize}
    \item[(i)] Compared with CMF and CV models, the proposed algorithm is more suitable for bone segmentation and can extract complete bone masks. 
    \item[(ii)] The U-Net trained on pelvic dataset achieves comparable performance for pelvic images, but fails to generalize to the ankle. Even though it is well known that the performance of deep learning methods depends strongly on the training dataset, the proposed model-based algorithm still shows advantages in cross-anatomy tasks and ensures a consistent performance.
    \item[(iii)] As for SAM, we are not aiming to make a decisive comparison. It has been witnessed that SAM holds surprising generality and ability in real-time processing. We also find that, just as SAM acknowledged, it can sometimes miss small features and produce vague boundaries, as shown in the second row of figure \ref{fig7}, which might not be tolerable for high precision tasks.
\end{itemize}

\begin{figure}[htbp]
    \centering
    \includegraphics[width=\textwidth]{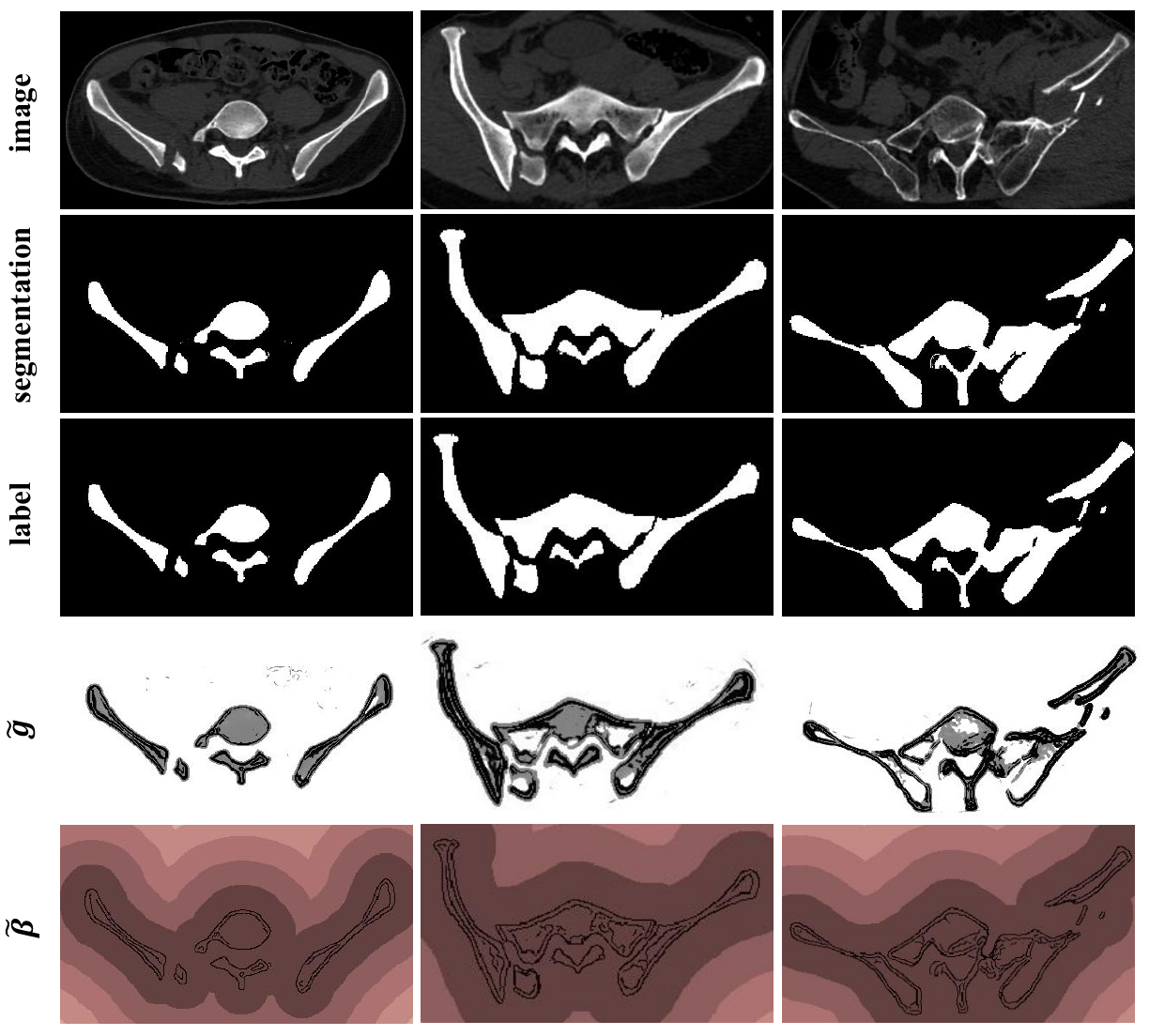} 
    \vspace{-3mm}
    \caption{More pelvic examples of the proposed algorithm.}
    \label{fig8}
\end{figure}

\begin{figure}[htbp]
    \centering
    \includegraphics[width=\textwidth]{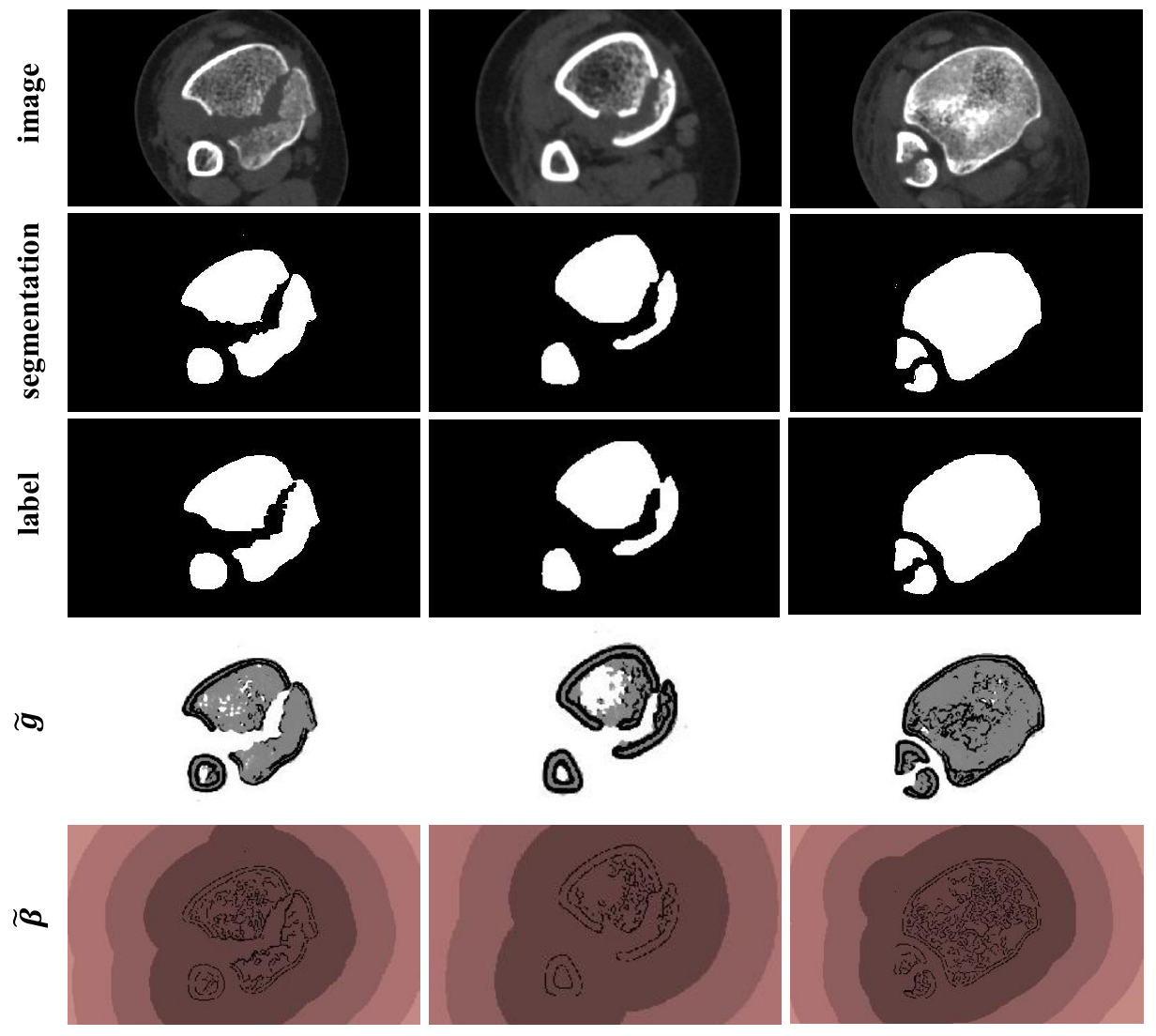}
        \vspace{-3mm}
    \caption{More ankle examples of the proposed algorithm.}
    \label{fig9}
\end{figure}

 We test these algorithms on two volumes in each dataset (independent of the training data for U-Net) and the corresponding average metrics are shown in table \ref{tab1}. For each comparison, score in \textbf{bold} means the best among tested methods. More results from the proposed algorithm are shown in figure \ref{fig8} and figure \ref{fig9}.

\begin{table}[htbp]
\centering
\caption{Quantitative comparisons between different methods.}
\label{tab1}
\vspace{5mm}
\begin{tabular}{cccccc}
\hline
Dataset                      & Methods  & Dice\,$\uparrow$           & Jaccard\,$\uparrow$       & HD\,$\downarrow$            & ASSD\,$\downarrow$         \\ \hline
\multirow{4}{*}{Pelvic (i)}  & CMF      & 45.70          & 29.73          & 14.06         & 1.85          \\
                             & CV       & 74.72          & 59.80          & 15.50         & 2.01          \\
                             & U-Net    & 94.74          & 91.84          & 6.75          & 0.61          \\
                             & proposed & \textbf{96.73} & \textbf{93.68} & \textbf{3.80} & \textbf{0.39} \\ \hline
\multirow{4}{*}{Pelvic (ii)} & CMF      & 51.02          & 34.36          & 14.18         & 2.06          \\
                             & CV       & 79.25          & 65.80           & 15.36         & 1.98          \\
                             & U-Net    & 96.44          & 93.13          & 3.74          & 0.51          \\
                             & proposed & \textbf{97.43} & \textbf{95.01} & \textbf{3.15} & \textbf{0.34} \\ \hline
\multirow{4}{*}{Ankle (iii)} & CMF      & 62.82          & 46.83          & 30.32         & 4.59          \\
                             & CV       & 75.94          & 63.68          & 25.17         & 3.68          \\
                             & U-Net    & 64.69          & 48.11          & 35.34         & 8.37          \\
                             & proposed & \textbf{97.80} & \textbf{95.69} & \textbf{5.15} & \textbf{0.52} \\ \hline
\end{tabular}
\end{table}

\subsection{Discussion}
\begin{figure}[htbp]
    \centering
    \includegraphics[width=\textwidth]{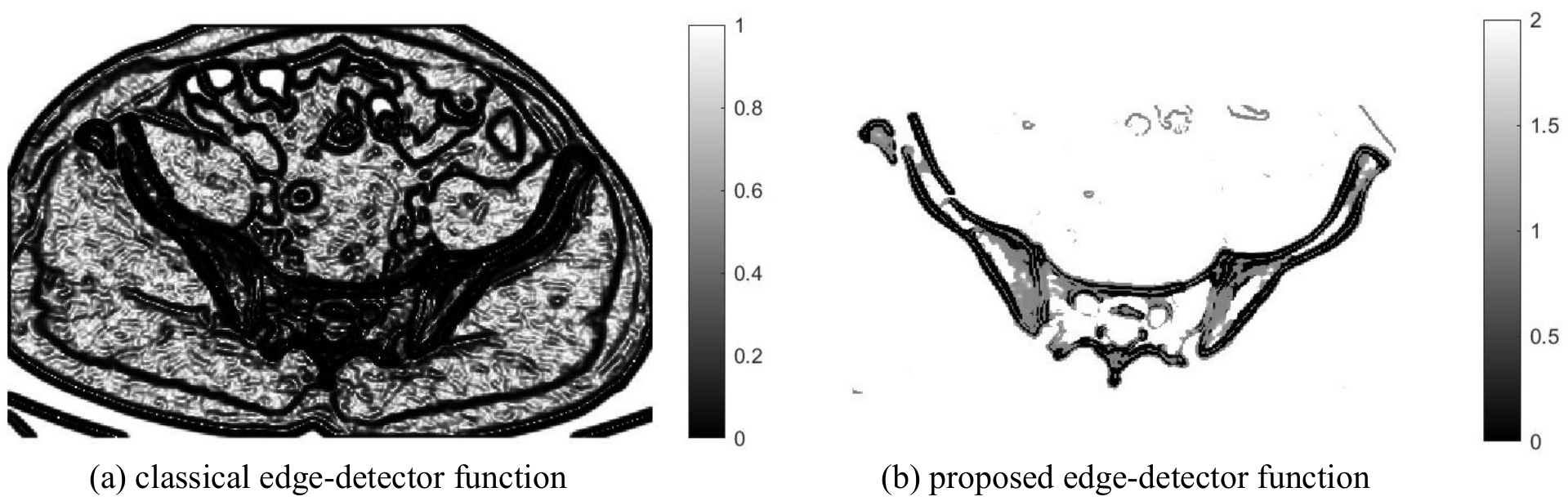}
\vspace{-3mm}
    \caption{A comparison between the classical and the proposed edge-detector functions. Figure 10(a) is the classical one, where the CT image is denoised by a $9\times 9$ Gaussian filter with standard deviation of 1.5 before the gradient operator. Figure 10(b) is the proposed one.}
    \label{fig10}
\end{figure}

Several parameters are involved in the proposed algorithm, and an extra elaboration should be laid on parameter tuning. Before that, a visual comparison between the classical edge-detector function in equation \eqref{J1} and the proposed one in equation \eqref{EDGE-DETECTOR} is illustrated in figure \ref{fig10}.
Obviously, the classical edge-detector function is blurry, where the information about the target is entangled with other objects, while in the proposed one the target is well extracted and highlighted. As for the proposed distance information, a perfect extraction of bone edges ensures a perfect stopping criterion. From these aspects, the proposed edge-detector function and distance information play the most significant roles in the proposed algorithm, since they extract bone features in different ways and guide the contour where to go and where to stop.

Basically, a referable principle of parameter tuning goes to obtain cleaner edge-detector function and distance information. Thus, once the concerned bone anatomy is determined, only a few parameters are worth adjusting: the parameters $\varepsilon_1,\varepsilon_2$ should be adjusted for the purpose of a cleaner edge-detector function, $\varepsilon_1$ determines how many bone intensity features are left and $\varepsilon_2$ determines how many weak edges are filtered; the double thresholds in the Canny algorithm and the threshold $\eta$ in \eqref{Ebone} should be adjusted for a better distance information, the former determines how fine the edges are extracted and the latter determines which edges are left as bone edges according to the averaged intensity in a small neighbourhood. As for the coefficient $\alpha$ of the area term, no particular adjustment is needed because the experiments in the subsection \ref{ablation} and figure \ref{fig6} testified the robustness with respect to $\alpha$, and the distance information $\Tilde{\beta}$ ensures the stopping information, while this robustness is not true for the classical GAC algorithm: as shown in figure \ref{fig5}, the contour collapses in very late iterations when $\alpha=1.0$ for the classical algorithm.

\section{Conclusion}
\label{Section 5}

Accurate segmentation of bone structures has been an important and challenging task because of its physiological properties. To overcome the challenge, this work aimed to provide an improved GAC algorithm for bone segmentation, which is intrinsically robust to bone fractures and soft tissue edges without depending on the annotated data. 

Our study found that the main limitation of the classical GAC algorithm is the indiscriminate edge extraction, which traps the model in the phenomena of edge obstruction, edge leakage and bone fracture, and inhibits the performance in bone segmentation. To address these problems point by point, we firstly proposed a novel edge-detector function which captures clear bone features and filters out other organs by orthopedic domain knowledge. It guides the contour towards bone edges without being obstructed and therefore reduces mis-segmentation. Next, we introduced a normalized distance map with respect to the bone edges as an adaptive step size to pace the contour evolution and help the contour stop at the bone edges, which makes the evolution more stable. Based on that, we further embedded fracture prompts into the distance information to help the contour stop on the fractures with very weak edges even without edges. This embedding provides a way to interact with fractures and improves the accuracy in fracture regions. The proposed algorithm emphasizes on the role of the proposed edge-detector function and the distance information. The two functions combine together to extract discriminative edge features and yield accurate, stable and consistent results. We compared the performance between the proposed algorithm and the classical algorithm, and tested the robustness of the proposed algorithm with respect to parameters, which demonstrate an efficient and robust performance on solving the aforementioned problems. More comparisons with other methods including SAM, U-Net, CMF and CV on both pelvic and ankle images suggest an accurate and consistent segmentation and broader applications to other bone anatomies. 

Admittedly, there are several limitations for the proposed algorithm, which deserve consideration in the future work. First of all, the performance of the proposed algorithm depends on the quality of the fracture prompts and the pre-defined CT windows or thresholds, which relies on the professionality of the physicians. Moreover, the proposed algorithm currently benefits fractures with relatively large bone fragments. For small fragments or complex bone structures with fine details, fracture prompts are not easy to provide. On the other hand, our work might also offer insights for researchers into combining domain knowledge and deep neural networks, such as predicting a signed distance function by neural networks and reformulating the proposed energy functional as a loss function, or learning a better edge-detector function directly from big data. The future work will be focused on automatic multi-class segmentation with sufficient exploitation of prior knowledge and mathematical principles.


\bibliographystyle{plain}
\bibliography{reference}

\begin{thebibliography}{10}

\bibitem{MATHEMATICAL_PROBLEMS_IN_IMAGE_PROCESSING}
G.~Aubert and P.~Kornprobst.
\newblock {\em Mathematical problems in image processing: partial differential
  equations and the calculus of variations}, volume 147.
\newblock Springer, New York, 2006.

\bibitem{IMAGING_OF_BONES_AND_JOINTS}
K.~Bohndorf, M.~W. Anderson, A.~M. Davies, H.~Imhof, and K.~Woertler.
\newblock {\em {Imaging of Bones and Joints: A Concise, Multimodality
  Approach}}.
\newblock Georg Thieme Verlag, 2016.

\bibitem{Cao2021SwinUnetUP}
H.~Cao, Y.~Wang, J.~Chen, D.~Jiang, X.~Zhang, Q.~Tian, and M.~Wang.
\newblock {Swin-Unet: Unet-like Pure Transformer for Medical Image
  Segmentation}.
\newblock In {\em ECCV Workshops}, pages 205--218, 2021.

\bibitem{GAC}
V.~Caselles, R.~Kimmel, and G.~Sapiro.
\newblock {Geodesic Active Contours}.
\newblock In {\em Proceedings of IEEE International Conference on Computer
  Vision}, pages 694--699, 1995.

\bibitem{GAC_REVIEW}
V.~Caselles, R.~Kimmel, and G.~Sapiro.
\newblock {Geometric Active Contours for Image Segmentation}.
\newblock In {\em Handbook of Image and Video Processing (Second Edition)},
  Communications, Networking and Multimedia, pages 613--627. Academic Press,
  second edition edition, 2005.

\bibitem{CHAN_VESE}
T.~Chan and L.~Vese.
\newblock {Active Contours without Edges}.
\newblock {\em IEEE Transactions on Image Processing}, 10(2):266--277, 2001.

\bibitem{PIECEWISE_POLYNOMIAL}
C.~Chen, J.~Leng, and G.~Xu.
\newblock {A General Framework of Piecewise-polynomial Mumford--Shah Model for
  Image Segmentation}.
\newblock {\em International Journal of Computer Mathematics},
  94(10):1981--1997, 2017.

\bibitem{Chen2017}
D.~Chen, J.-M. Mirebeau, and L.~Cohen.
\newblock Global minimum for a {F}insler {E}lastica minimal path approach.
\newblock {\em International Journal of Computer Vision}, 122(3):458--483,
  2017.

\bibitem{Chen2024}
D.~Chen, J.-M. Mirebeau, H.~Shu, and L.~Cohen.
\newblock {A Region-Based {R}anders Geodesic Approach for Image Segmentation}.
\newblock {\em International Journal of Computer Vision}, 132(2):349--391,
  2024.

\bibitem{CHEN2023109728}
G.~Chen, L.~Li, J.~Zhang, and Y.~Dai.
\newblock {Rethinking the unpretentious U-net for medical ultrasound image
  segmentation}.
\newblock {\em Pattern Recognition}, 142:109728, 2023.

\bibitem{3DUNET}
{\"O}.~{\c{C}}i{\c{c}}ek, A.~Abdulkadir, S.~S. Lienkamp, T.~Brox, and
  O.~Ronneberger.
\newblock {3D U-Net: Learning Dense Volumetric Segmentation from Sparse
  Annotation}.
\newblock In {\em Medical Image Computing and Computer-Assisted Intervention --
  MICCAI 2016}, pages 424--432, 2016.

\bibitem{Cohen1997}
L.~Cohen and R.~Kimmel.
\newblock {Global Minimum for Active Contour Models: A Minimal Path Approach}.
\newblock {\em International Journal of Computer Vision}, 24(1):57--78, 1997.

\bibitem{Duits2018}
R.~Duits, S.~Meesters, J.-M. Mirebeau, and J.~Portegies.
\newblock {Optimal Paths for Variants of the {2D} and {3D} {R}eeds--{S}hepp Car
  with Applications in Image Analysis}.
\newblock {\em Journal of Mathematical Imaging and Vision}, 60(6):816--848,
  2018.

\bibitem{GANGWAR201895}
T.~Gangwar, J.~Calder, T.~Takahashi, J.~E. Bechtold, and D.~Schillinger.
\newblock {Robust variational segmentation of 3D bone CT data with thin
  cartilage interfaces}.
\newblock {\em Medical Image Analysis}, 47:95--110, 2018.

\bibitem{9413346}
C.~Guo, M.~Szemenyei, Y.~Yi, W.~Wang, B.~Chen, and C.~Fan.
\newblock {SA-UNet: Spatial Attention U-Net for Retinal Vessel Segmentation}.
\newblock In {\em 2020 25th International Conference on Pattern Recognition
  (ICPR)}, pages 1236--1242, 2021.

\bibitem{HAN2020107520}
B.~Han and Y.~Wu.
\newblock {Active contour model for inhomogenous image segmentation based on
  Jeffreys divergence}.
\newblock {\em Pattern Recognition}, 107:107520, 2020.

\bibitem{han2021fracture}
R.~Han, A.~Uneri, R.~C. Vijayan, P.~Wu, P.~Vagdargi, N.~Sheth, S.~Vogt,
  G.~Kleinszig, G.~M. Osgood, and J.~H. Siewerdsen.
\newblock Fracture reduction planning and guidance in orthopaedic trauma
  surgery via multi-body image registration.
\newblock {\em Medical Image Analysis}, 68:101917, 2021.

\bibitem{NNUNET}
F.~Isensee, P.~F. Jaeger, S.~A.~A. Kohl, J.~Petersen, and K.~Maier-Hein.
\newblock {nnU-Net: A Self-configuring Method for Deep Learning-based
  Biomedical Image Segmentation}.
\newblock {\em Nature Methods}, 18:203--211, 2020.

\bibitem{1663773}
Y.~Jia and Y.~Jiang.
\newblock {Active Contour Model with Shape Constraints for Bone Fracture
  Detection}.
\newblock In {\em International Conference on Computer Graphics, Imaging and
  Visualisation (CGIV'06)}, pages 90--95, 2006.

\bibitem{SAM}
A.~Kirillov, E.~Mintun, N.~Ravi, H.~Mao, C.~Rolland, L.~Gustafson, T.~Xiao,
  S.~Whitehead, A.~C. Berg, W.-Y. Lo, P.~Doll{\'a}r, and R.~B. Girshick.
\newblock {Segment Anything}.
\newblock {\em 2023 IEEE/CVF International Conference on Computer Vision
  (ICCV)}, pages 3992--4003, 2023.

\bibitem{KORFIATIS2017358}
V.~Korfiatis, S.~Tassani, and G.~K. Matsopoulos.
\newblock {An Independent Active Contours Segmentation framework for bone
  micro-CT images}.
\newblock {\em Computers in Biology and Medicine}, 87:358--370, 2017.

\bibitem{RSF}
C.~Li, C.-Y. Kao, J.~C. Gore, and Z.~Ding.
\newblock {Minimization of Region-Scalable Fitting Energy for Image
  Segmentation}.
\newblock {\em IEEE Transactions on Image Processing}, 17(10):1940--1949, 2008.

\bibitem{DRLSE}
C.~Li, C.~Xu, C.~Gui, and M.~D. Fox.
\newblock {Distance Regularized Level Set Evolution and Its Application to
  Image Segmentation}.
\newblock {\em IEEE Transactions on Image Processing}, 19(12):3243--3254, 2010.

\bibitem{li2018h}
X.~Li, H.~Chen, X.~Qi, Q.~Dou, C.-W. Fu, and P.-A. Heng.
\newblock {H-DenseUNet: hybrid densely connected UNet for liver and tumor
  segmentation from CT volumes}.
\newblock {\em IEEE Transactions on Medical Imaging}, 37(12):2663--2674, 2018.

\bibitem{DEEP_LEARNING_TO_SEGMENT_PELVIC}
P.~Liu, H.~Han, Du~Y., H.~Zhu, Y.~Li, F.~Gu, H.~Xiao, J.~Li, C.~Zhao, L.~Xiao,
  X.~Wu, and S.~Zhou.
\newblock {Deep Learning to Segment Pelvic Bones: Large-scale CT Datasets and
  Baseline Models}.
\newblock {\em International Journal of Computer Assisted Radiology and
  Surgery}, 16:749--756, 2020.

\bibitem{9815506}
W.~Liu, H.~Yang, T.~Tian, Z.~Cao, X.~Pan, W.~Xu, Y.~Jin, and F.~Gao.
\newblock {Full-Resolution Network and Dual-Threshold Iteration for Retinal
  Vessel and Coronary Angiograph Segmentation}.
\newblock {\em IEEE Journal of Biomedical and Health Informatics},
  26(9):4623--4634, 2022.

\bibitem{PELVIC_FRACTURE_SEGMENTATION}
Y.~Liu, S.~Yibulayimu, Y.~Sang, G.~Zhu, Y.~Wang, C.~Zhao, and X.~Wu.
\newblock {Pelvic Fracture Segmentation Using a Multi-scale Distance-Weighted
  Neural Network}.
\newblock In {\em Medical Image Computing and Computer Assisted Intervention --
  MICCAI 2023}, pages 312--321, 2023.

\bibitem{MEDSAM}
J.~Ma, Y.~He, F.~Li, L.-J. Han, C.~You, and B.~Wang.
\newblock {Segment Anything in Medical Images}.
\newblock {\em Nature Communications}, 15, 2023.

\bibitem{CT_WINDOWS2}
J.~C. Mandell, J.~R. Wortman, T.~C. Rocha, L.~R. Folio, K.~P. Andriole, and
  B.~Khurana.
\newblock {Computed Tomography Window Blending: Feasibility in Thoracic
  Trauma}.
\newblock {\em Academic radiology}, 25(9):1190--1200, 2018.

\bibitem{FAC}
J.~Melonakos, E.~Pichon, S.~Angenent, and A.~Tannenbaum.
\newblock {Finsler Active Contours}.
\newblock {\em IEEE Transactions on Pattern Analysis and Machine Intelligence},
  30(3):412--423, 2008.

\bibitem{VNET}
F.~Milletari, N.~Navab, and S.-A. Ahmadi.
\newblock {V-Net: Fully Convolutional Neural Networks for Volumetric Medical
  Image Segmentation}.
\newblock In {\em 2016 Fourth International Conference on 3D Vision (3DV)},
  pages 565--571, 2016.

\bibitem{MIN201969}
H.~Min, L.~Xia, J.~Han, X.~Wang, Q.~Pan, H.~Fu, H.~Wang, S.~Wong, and H.~Li.
\newblock A multi-scale level set method based on local features for
  segmentation of images with intensity inhomogeneity.
\newblock {\em Pattern Recognition}, 91:69--85, 2019.

\bibitem{UNET}
O.~Ronneberger, P.~Fischer, and T.~Brox.
\newblock {U-Net: Convolutional Networks for Biomedical Image Segmentation}.
\newblock In {\em Medical Image Computing and Computer-Assisted Intervention --
  MICCAI 2015}, pages 234--241, 2015.

\bibitem{SHAN2024110007}
H.~Shan.
\newblock {CS-GAC: Compressively sensed geodesic active contours}.
\newblock {\em Pattern Recognition}, 146:110007, 2024.

\bibitem{4906670}
S.~Vasilache and K.~Najarian.
\newblock {A unified method based on wavelet filtering and Active Contour
  Models for segmentation of Pelvic CT images}.
\newblock In {\em 2009 ICME International Conference on Complex Medical
  Engineering}, pages 1--5, 2009.

\bibitem{TRANSFORMER}
A.~Vaswani, N.~M. Shazeer, N.~Parmar, J.~Uszkoreit, L.~Jones, A.~N. Gomez,
  L.~Kaiser, and I.~Polosukhin.
\newblock {Attention is All You Need}.
\newblock In {\em Neural Information Processing Systems}, pages 6000--6010,
  2017.

\bibitem{ICTM}
D.~Wang and X.-P. Wang.
\newblock {The Iterative Convolution–Thresholding Method (ICTM) for Image
  Segmentation}.
\newblock {\em Pattern Recognition}, 130:108794, 2022.

\bibitem{YANG2021107985}
Y.~Yang, R.~Wang, X.~Shu, C.~Feng, R.~Xie, W.~Jia, and C.~Li.
\newblock Level set framework with transcendental constraint for robust and
  fast image segmentation.
\newblock {\em Pattern Recognition}, 117:107985, 2021.

\bibitem{CT_WINDOWS1}
J.~Yu, Z.~Li, and J.~Gao.
\newblock {Technical Specifications for CT Examination of Severe Traumatic
  Injuries}.
\newblock {\em Chinese Journal of Radiology}, 38(21):12, 2023.

\bibitem{CONTINUOUS_MAX_FLOW_AND_MIN_CUT}
J.~Yuan, E.~Bae, X.-C. Tai, and Y.~Boykov.
\newblock {A Spatially Continuous Max-flow and Min-cut Framework for Binary
  Labeling Problems}.
\newblock {\em Numerische Mathematik}, 126:559--587, 2014.

\bibitem{RESUNET}
Z.~Zhang, Q.~Liu, and Y.~Wang.
\newblock {Road Extraction by Deep Residual U-Net}.
\newblock {\em IEEE Geoscience and Remote Sensing Letters}, 15(5):749--753,
  2018.

\bibitem{A_VARIATIONAL_LEVEL_SET_APPROACH}
H.~Zhao, T.~Chan, B.~Merriman, and S.~Osher.
\newblock {A Variational Level Set Approach to Multiphase Motion}.
\newblock {\em Journal of Computational Physics}, 127(1):179--195, 1996.

\end{thebibliography}

\end{document}